\documentclass{article}
\usepackage{spconf,amsmath,graphicx}
\usepackage[hidelinks]{hyperref}
\usepackage{cite}
\usepackage{amssymb}
\usepackage{booktabs}
\usepackage[table]{xcolor}

\newcommand{\method}{Endo-TSR}

\title{ENDO-TSR: TEMPORAL SPECTRAL MODELING OF APPEARANCE AND MOTION FOR ENDOSCOPIC RECONSTRUCTION}
\name{Taoyu Wu$^{1,3}$
\quad Yiyi Miao$^{1,3}$
\quad Qi Shao$^{3}$
\quad Zhuoxiao Li$^{4}$
\quad Zhe Tang$^{2}$
\quad Limin Yu$^{1}$
\quad Baoru Huang$^{3}$}
\address{{\small $^{1}$ Academy of Artificial Intelligence and Advanced Technology, Xi'an Jiaotong-Liverpool University, China}\\
{\small $^{2}$ Institute of Artificial Intelligence Innovation, Zhejiang University of Technology, China}\\
{\small $^{3}$ School of Computer Science and Informatics, University of Liverpool, United Kingdom}\\
{\small $^{4}$ Hong Kong University of Science and Technology (Guangzhou), China}}
\begin{document}
\ninept
\maketitle
\begin{abstract}
Endoscopic scene reconstruction requires modeling tissue motion and temporal appearance while recovering fine surface detail.
Deformable Gaussian models provide explicit trajectories, but their fixed colour coefficients lack a dedicated temporal representation for photometric changes.
We propose \method{}, which augments deformable Gaussian splatting with bounded Fourier colour residuals and independent translation residuals on shared temporal frequencies.
The colour residuals capture local appearance changes, while a Mat\'ern spectral prior regularises motion corrections.
Multi-scale Laplacian supervision guides tissue-detail recovery during joint image fitting.
Extensive experiments on the EndoNeRF and StereoMIS datasets demonstrate state-of-the-art rendering quality, with the highest PSNR across all evaluated sequences.
Ablation studies show that temporal appearance yields the largest PSNR gain among the tested component additions, while appearance and detail supervision jointly improve rendering with fixed Gaussian counts.
\end{abstract}
\begin{keywords}
Endoscopic scene reconstruction, dynamic Gaussian splatting, temporal appearance, spectral regularisation, Mat\'ern kernel
\end{keywords}
\section{Introduction}
\label{sec:intro}

Endoscopic reconstruction seeks a dynamic tissue representation for surgical navigation, robotic assistance and simulation~\cite{yang20243d,wang2022neuralendonerf}.
As tissue deforms, the nearby endoscope light changes shading and reflections on wet surfaces.
Reproducing the observed scene therefore requires both temporal appearance and motion modeling, together with the recovery of tissue detail.

Neural radiance and surface representations support dynamic endoscopic reconstruction~\cite{mildenhall2021nerf,wang2022neuralendonerf,zha2023endosurf}, with space--time factorisation improving efficiency~\cite{yang2024forplane}.
Gaussian splatting enables differentiable rendering of explicit primitives~\cite{3dgs}, with deformation extending this representation to dynamic scenes~\cite{wu20244d}.
Endoscopic methods strengthen this representation through depth guidance~\cite{liu2024endogaussian,huang2024endo4dgs}, flexible tissue trajectories~\cite{yang2024deform3dgs,chen2024surgicalgs} and pair priors for spatial initialisation~\cite{yu2026endopairgs}.
Related advances include flow-constrained joint mapping and pose estimation~\cite{wu2025endoflowslam} and phase-adaptive appearance with rational-wavelet supervision~\cite{wu2025endowave}.
Deformation-based formulations with fixed colour coefficients assign temporal variation to geometry~\cite{yang2024deform3dgs,chen2024surgicalgs}.
View-dependent colour describes directional variation but lacks an explicit temporal variable for tissue that brightens or darkens as illumination changes.

We introduce \method{} to give temporal appearance its own representation alongside tissue motion.
We capture local photometric variation with a bounded Fourier colour residual and refine each Gaussian's trajectory with an independent translation residual.
The residuals share temporal frequencies but have distinct roles and coefficients.
A Mat\'ern spectral prior assigns stronger penalties to higher motion frequencies, favouring smooth trajectory corrections.
To guide the recovery of spatial detail, a multi-scale Laplacian objective supervises tissue texture and boundaries in the rendered images.
Together, these components jointly fit appearance and motion to image evidence, with separate constraints on motion frequencies and spatial detail.

Our contributions are:
\textbf{(i)} a bounded temporal appearance representation for deformable endoscopic Gaussian splatting.
\textbf{(ii)} a complementary refinement formulation combining independent colour and translation residuals, a motion spectral prior and multi-scale detail supervision.
\textbf{(iii)} extensive experiments on EndoNeRF and StereoMIS demonstrating state-of-the-art rendering quality, with ablations establishing the gains from temporal appearance and detail supervision, including their joint benefit at fixed Gaussian counts.

\begin{figure*}[t]
    \centering
    \includegraphics[width=\textwidth]{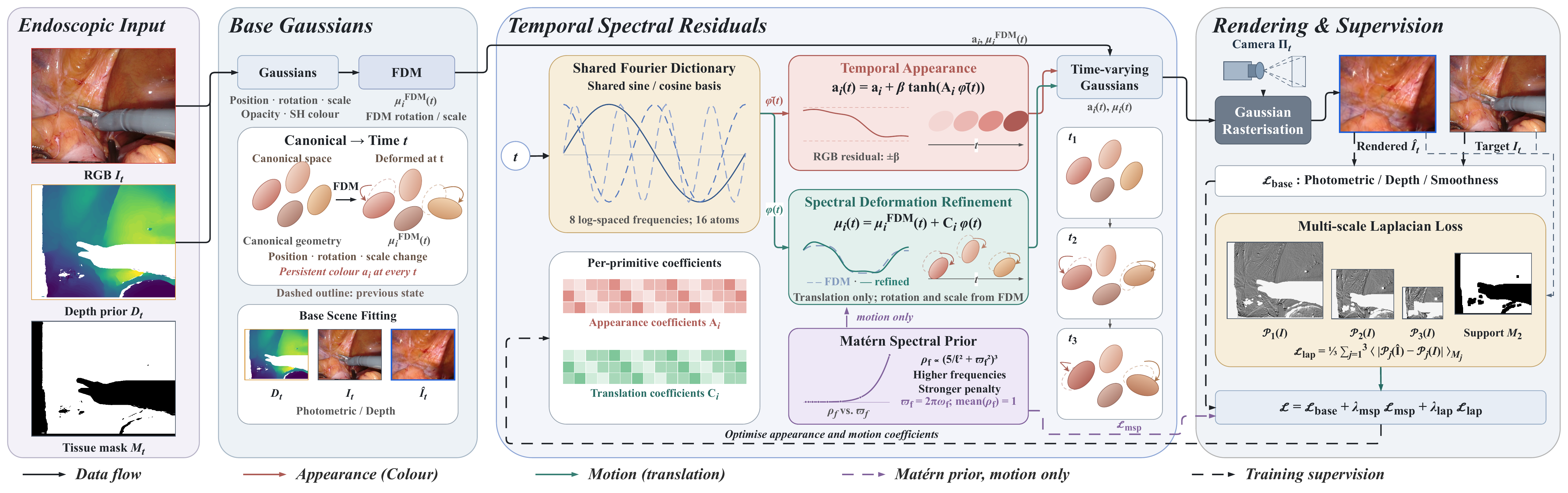}
    \caption{Overview of \method{}. RGB images and depth priors establish base Gaussians, whose persistent colour and FDM trajectories form the basis of the time-varying representation. In parallel, a shared Fourier dictionary and independent coefficients produce bounded colour residuals and translation corrections. These residuals augment the base quantities before rasterisation at the input camera pose. Training combines base reconstruction losses, multi-scale Laplacian supervision over valid tissue regions, and a Mat\'ern penalty on motion coefficients. Dashed Gaussian outlines indicate previous states. Solid arrows denote forward flow. Black dashed arrows return supervision to the residual coefficients, while purple dashed arrows connect the motion prior to the translation branch and total objective.}
    \label{fig:pipeline}
\end{figure*}

\section{Method}
\label{sec:method}

We reconstruct a dynamic Gaussian scene from an endoscopic sequence $\{I_t,M_t,D_t,\Pi_t\}$, where $I_t$ is an RGB image, $M_t$ is a tissue mask, $D_t$ is a depth prior, and $\Pi_t$ contains camera parameters.
The representation supports rendering at observed camera poses and intermediate times.
We initialise $N$ Gaussians from training views and model their trajectories with a flexible deformation model (FDM) under depth supervision~\cite{chen2024surgicalgs}.
Each Gaussian stores a centre, rotation, scale, opacity and spherical-harmonic (SH) colour coefficients.
The FDM deforms each Gaussian's canonical geometry at time $t$, providing its base trajectory for reconstruction.

\method{} represents temporal variation through separate appearance and motion variables (Fig.~\ref{fig:pipeline}).
A bounded colour residual updates constant SH coefficients, while a translation residual corrects Gaussian centres.
The residuals use shared temporal frequencies and are applied before differentiable rasterisation.
The rendered image couples their optimisation, guided by a spectral prior on motion and multi-scale tissue-detail supervision.
We initialise the temporal coefficients to zero and activate them at iteration $K_s$ after initial scene fitting.

\subsection{Temporal appearance modeling}
\label{sec:appearance}

The colour residual allows each Gaussian to follow local photometric changes as illumination evolves.
Let $\mathbf{a}_i\in\mathbb{R}^{3}$ be its persistent view-independent RGB colour, corresponding to the degree-zero SH coefficient.
We express temporal variation with a Fourier dictionary~\cite{tancik2020fourier}.
For normalised time $t\in[0,1]$,
\begin{equation}
\begin{split}
\phi(t) = [&\sin(2\pi\omega_1t),\ldots,\sin(2\pi\omega_Ft),\\
           &\cos(2\pi\omega_1t),\ldots,\cos(2\pi\omega_Ft)]^{\top},
\end{split}
\label{eq:dict}
\end{equation}
where $\omega_f$ are $F$ logarithmically spaced frequencies.
For appearance modeling, we centre each atom over the training time stamps and normalise it to unit root-mean-square (RMS) magnitude, obtaining $\bar\phi(t)$.
Centring removes the temporal mean of each input atom, while normalisation places the basis atoms on a common scale.
We reuse these training-time statistics at held-out times.
The coefficients are shared across the sequence, so rendering at new times queries the same continuous basis without frame-specific colour parameters.

The time-dependent colour is the persistent colour plus a bounded residual:
\begin{equation}
\mathbf{a}_i(t)=\mathbf{a}_i+\beta\tanh\!\big(\mathbf{A}_i\bar\phi(t)\big),
\label{eq:appearance}
\end{equation}
where $\mathbf{A}_i\in\mathbb{R}^{3\times2F}$ contains per-Gaussian, per-channel coefficients.
These coefficients describe local appearance changes, while the hyperbolic tangent bounds each RGB correction by $\beta$.
We convert the residual to SH units as $\Delta\mathbf{h}_i^{0}(t)=[\mathbf{a}_i(t)-\mathbf{a}_i]/Y_{00}$, where $Y_{00}=1/\sqrt{4\pi}$, and add it to the base coefficient before rendering.
Higher-order SH coefficients continue to model view dependence.
Initialising $\mathbf{A}_i=0$ leaves colour unchanged at activation, after which image gradients jointly refine appearance and geometry.

\begin{table*}[t]
\caption{\textbf{Rendering quality on EndoNeRF and StereoMIS} under the Endo-PairGS benchmark protocol~\cite{yu2026endopairgs}, with full-image averaging. Our results are averaged over five seeds on EndoNeRF and three on StereoMIS. Bold: best per column.}
\centering
{\scriptsize
\setlength{\tabcolsep}{1.8pt}
\renewcommand{\arraystretch}{1.38}
\begin{tabular}{@{}l ccc ccc ccc ccc ccc ccc@{}}
\toprule
& \multicolumn{3}{c}{EndoNeRF--Cutting} & \multicolumn{3}{c}{EndoNeRF--Pulling} & \multicolumn{3}{c}{StereoMIS--P1} & \multicolumn{3}{c}{StereoMIS--P2-2} & \multicolumn{3}{c}{StereoMIS--P2-5} & \multicolumn{3}{c}{StereoMIS--P3} \\
\cmidrule(lr){2-4}\cmidrule(lr){5-7}\cmidrule(lr){8-10}\cmidrule(lr){11-13}\cmidrule(lr){14-16}\cmidrule(lr){17-19}
Method & PSNR$\uparrow$ & SSIM$\uparrow$ & LPIPS$\downarrow$ & PSNR$\uparrow$ & SSIM$\uparrow$ & LPIPS$\downarrow$ & PSNR$\uparrow$ & SSIM$\uparrow$ & LPIPS$\downarrow$ & PSNR$\uparrow$ & SSIM$\uparrow$ & LPIPS$\downarrow$ & PSNR$\uparrow$ & SSIM$\uparrow$ & LPIPS$\downarrow$ & PSNR$\uparrow$ & SSIM$\uparrow$ & LPIPS$\downarrow$ \\
\midrule
EndoNeRF~\cite{wang2022neuralendonerf}  & 35.71 & 0.933 & 0.059 & 36.50 & 0.941 & 0.054 & 31.12 & 0.846 & 0.148 & 28.91 & 0.703 & 0.261 & 26.49 & 0.688 & 0.274 & 29.85 & 0.751 & 0.209 \\
EndoSurf~\cite{zha2023endosurf}         & 36.55 & 0.953 & 0.055 & 37.21 & 0.955 & 0.050 & 31.29 & 0.865 & 0.168 & 30.01 & 0.780 & 0.198 & 26.51 & 0.738 & 0.277 & 30.46 & 0.811 & 0.188 \\
ForPlane-32k~\cite{yang2024forplane}    & 36.85 & 0.944 & 0.045 & 37.82 & 0.955 & 0.042 & 32.19 & 0.843 & 0.143 & 28.01 & 0.659 & 0.286 & 28.48 & 0.759 & 0.170 & 31.10 & 0.801 & 0.139 \\
Endo-4DGS~\cite{huang2024endo4dgs}      & 36.56 & 0.955 & 0.032 & 37.85 & 0.959 & 0.043 & 32.17 & 0.850 & 0.192 & 31.45 & 0.812 & 0.141 & 29.06 & 0.817 & 0.152 & 31.21 & 0.831 & 0.155 \\
EndoGaussian~\cite{liu2024endogaussian} & 38.38 & 0.963 & 0.030 & 37.09 & 0.956 & 0.039 & 32.79 & 0.858 & 0.182 & 29.82 & 0.759 & 0.169 & 28.28 & 0.782 & 0.173 & 31.26 & 0.821 & 0.166 \\
Deform3DGS~\cite{yang2024deform3dgs}    & 38.72 & 0.966 & 0.030 & 38.33 & 0.960 & 0.040 & 33.35 & 0.871 & 0.169 & 30.69 & 0.789 & 0.144 & 29.16 & 0.813 & 0.147 & 31.70 & 0.834 & 0.160 \\
SurgicalGS~\cite{chen2024surgicalgs}    & 38.85 & 0.967 & 0.030 & 38.03 & 0.959 & 0.039 & 32.98 & 0.867 & 0.172 & 30.19 & 0.775 & 0.178 & 29.07 & 0.805 & 0.168 & 31.97 & 0.841 & 0.155 \\
Endo-PairGS~\cite{yu2026endopairgs}     & 38.33 & 0.967 & 0.020 & 38.96 & \textbf{0.967} & \textbf{0.022} & 33.56 & 0.871 & 0.146 & 32.16 & 0.833 & 0.100 & 29.75 & 0.846 & 0.103 & 32.47 & 0.871 & 0.110 \\
\midrule
\textbf{\method{} (Ours)} & \textbf{40.96} & \textbf{0.980} & \textbf{0.019} & \textbf{39.34} & \textbf{0.967} & 0.035 & \textbf{38.45} & \textbf{0.947} & \textbf{0.074} & \textbf{34.83} & \textbf{0.879} & \textbf{0.065} & \textbf{31.99} & \textbf{0.894} & \textbf{0.067} & \textbf{36.19} & \textbf{0.939} & \textbf{0.045} \\
\bottomrule
\end{tabular}}
\label{tab:sota}
\end{table*}

\subsection{Spectral motion refinement}
\label{sec:residual}
\label{sec:msp}

Motion refinement retains the FDM trajectory and adds a frequency-resolved correction to each Gaussian centre:
\begin{equation}
\mu_i(t)=\mu_i^{\mathrm{FDM}}(t)+\mathbf{C}_i\phi(t),
\label{eq:residual}
\end{equation}
where $\mu_i^{\mathrm{FDM}}(t)$ is the base centre and $\mathbf{C}_i\in\mathbb{R}^{3\times2F}$ holds zero-initialised translation coefficients.
Rotation and scale remain governed by the FDM~\cite{yang2024deform3dgs,chen2024surgicalgs}.
Motion and appearance share the frequencies in Eq.~\eqref{eq:dict}, but motion uses the unnormalised dictionary $\phi$, whereas appearance uses $\bar\phi$.
Their independent coefficients allow image fitting to adjust position and colour through distinct variables.

We regularise the translation coefficients to favour smooth trajectory corrections.
Bochner's spectral representation links a stationary covariance kernel to its spectral density~\cite{rasmussen2006gp}.
The Mat\'ern-5/2 density is $S(\varpi)\propto(5/\ell^2+\varpi^2)^{-3}$, where $\ell$ controls the temporal correlation scale and $\varpi_f=2\pi\omega_f$.
We use the inverse of this density to define the normalised weights and penalty
\begin{equation}
\begin{split}
\rho_f &= \frac{(5/\ell^2+\varpi_f^2)^3}{F^{-1}\sum_{g=1}^{F}(5/\ell^2+\varpi_g^2)^3},\\
\mathcal{L}_{\mathrm{msp}} &= \frac{1}{6NF}\sum_{i=1}^{N}\sum_{c=1}^{3}\sum_{f=1}^{F}
\rho_f\big(C_{i,c,f}^{2}+C_{i,c,F+f}^{2}\big).
\end{split}
\label{eq:msploss}
\end{equation}
The weights penalise higher motion frequencies more strongly.
Mean normalisation separates this relative frequency preference from the overall penalty strength.
For each frequency, the paired sine and cosine coefficients determine the amplitude and phase of the correction.
Equal weights on their squared magnitudes make the penalty phase-invariant: amplitude is penalised regardless of phase, retaining flexibility in when a trajectory correction occurs.
This prior acts on translation coefficients, while colour residuals receive image supervision.
GP-4DGS~\cite{kim2026gp4dgs} uses Gaussian processes for probabilistic motion prediction.
Here, the kernel instead defines a closed-form coefficient penalty during scene fitting.

\subsection{Multi-scale detail supervision}
\label{sec:laplacian}
\label{sec:curriculum}

The temporal residuals alter rendered appearance and position, so their supervision must also capture tissue texture and boundaries.
We use a Laplacian pyramid~\cite{burt1983laplacian} to compare this spatial detail across scales.
Starting from $G_0(I)=I$, we construct a Gaussian pyramid using binomial low-pass filtering and stride-two downsampling.
Each Laplacian level subtracts the bilinearly upsampled next Gaussian-pyramid level, $\mathcal{P}_j(I)=G_{j-1}(I)-\operatorname{up}(G_j(I))$.

For rendered and observed images $\hat I$ and $I$, we minimise
\begin{equation}
\mathcal{L}_{\mathrm{lap}}=\frac{1}{3}\sum_{j=1}^{3}
\Big\langle\big|\mathcal{P}_j(\hat I)-\mathcal{P}_j(I)\big|\Big\rangle_{M_j},
\label{eq:lap}
\end{equation}
where $\langle\cdot\rangle_{M_j}$ averages over RGB channels and valid pixels at level $j$.
The low-pass filter is a separable $5\times5$ binomial kernel.
We construct the pyramid before masking, avoiding artificial boundaries from zeroed instrument pixels.
The mask $M_j$ propagates validity through the same filtering and interpolation operations, retaining coefficients whose complete support lies within tissue.
This validity test is applied at each scale, so an instrument boundary cannot enter the detail objective through a neighbouring filter footprint.

Since appearance and motion jointly affect the rendered image, we refine them using a common training objective:
\begin{equation}
\mathcal{L}=\mathcal{L}_{\mathrm{base}}
+\lambda_{\mathrm{msp}}\mathcal{L}_{\mathrm{msp}}
+\lambda_{\mathrm{lap}}\mathcal{L}_{\mathrm{lap}},
\label{eq:total}
\end{equation}
where $\mathcal{L}_{\mathrm{base}}$ contains the base masked photometric, depth and smoothness losses.
Within this objective, the image losses jointly update colour and translation coefficients, depth constrains geometry, and the spectral penalty shapes the motion correction.

To set the strength of detail supervision relative to photometric fitting, we calibrate $\lambda_{\mathrm{lap}}$ at a base-model checkpoint over temporally distributed training views $\mathcal{V}$.
Let $g_q(v)=\operatorname{RMS}(\nabla_{\hat I_v}\mathcal{L}_q)$ for the photometric loss ($q=\mathrm{rgb}$) and detail loss ($q=\mathrm{lap}$).
For a target relative strength $r$, we use
\begin{equation}
\lambda_{\mathrm{lap}}=r\left[\operatorname{median}_{v\in\mathcal{V}}
\frac{g_{\mathrm{lap}}(v)}{g_{\mathrm{rgb}}(v)}\right]^{-1}.
\label{eq:calibration}
\end{equation}
Calibration compares gradients with respect to the rendered image, through which both residuals receive supervision.
The detail weight rises linearly during initial fitting, temporal refinement starts at $K_s$, and density adaptation stops before the final iterations.
The remaining iterations refine geometry and appearance with the Gaussian count fixed.

\begin{figure*}[t]
    \centering
    \begin{minipage}[t]{\dimexpr0.5\textwidth-0.5\columnsep\relax}
    \centering
    \includegraphics[width=\linewidth]{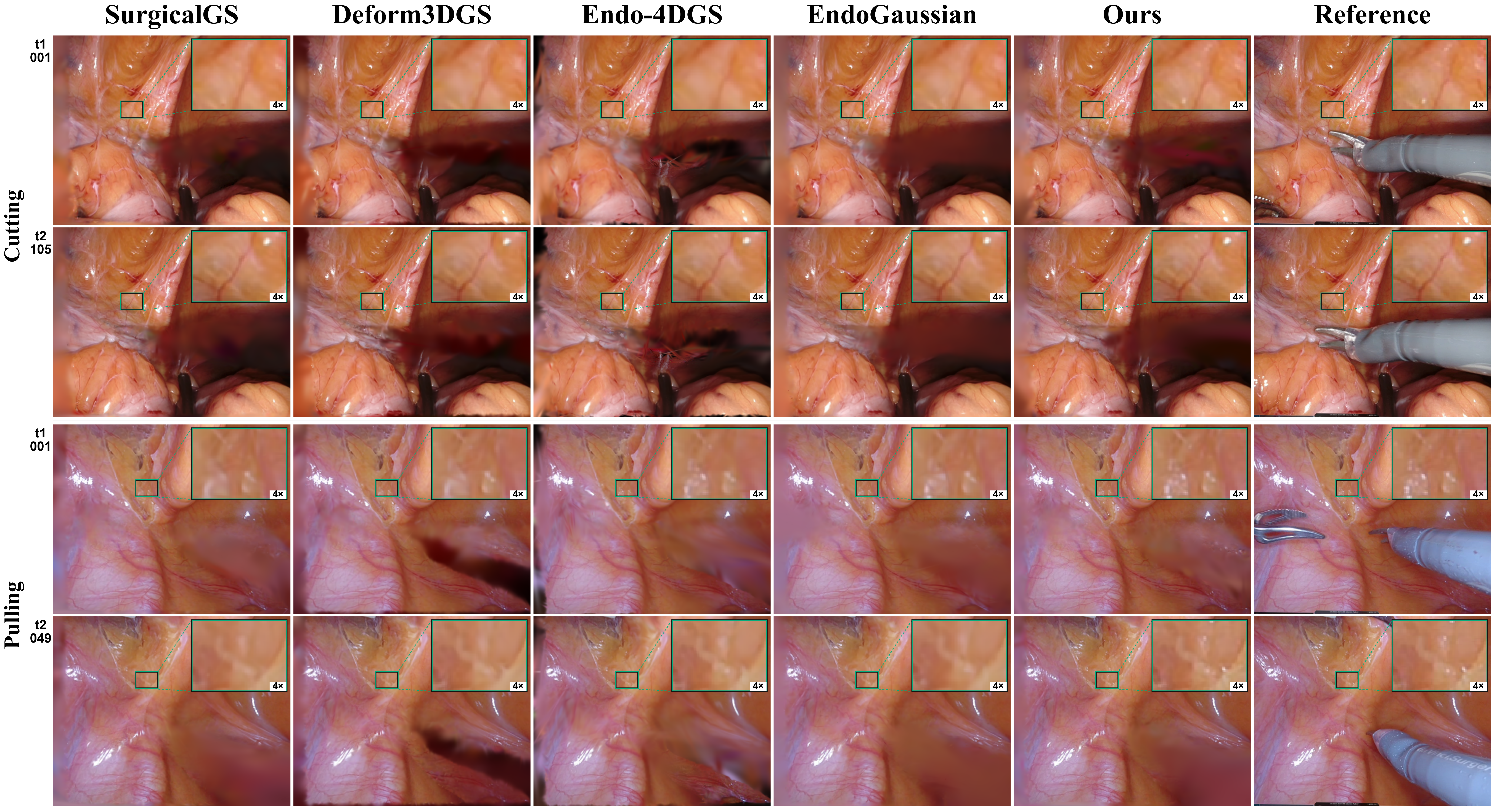}
    \caption{Held-out EndoNeRF frames at different times. The cutting (top) and pulling (bottom) examples show temporal appearance and tissue detail, with $4\times$ insets. \method{} preserves finer structure in the highlighted regions.}
    \label{fig:qual_endonerf}
    \end{minipage}\hfill
    \begin{minipage}[t]{\dimexpr0.5\textwidth-0.5\columnsep\relax}
    \centering
    \includegraphics[width=\linewidth]{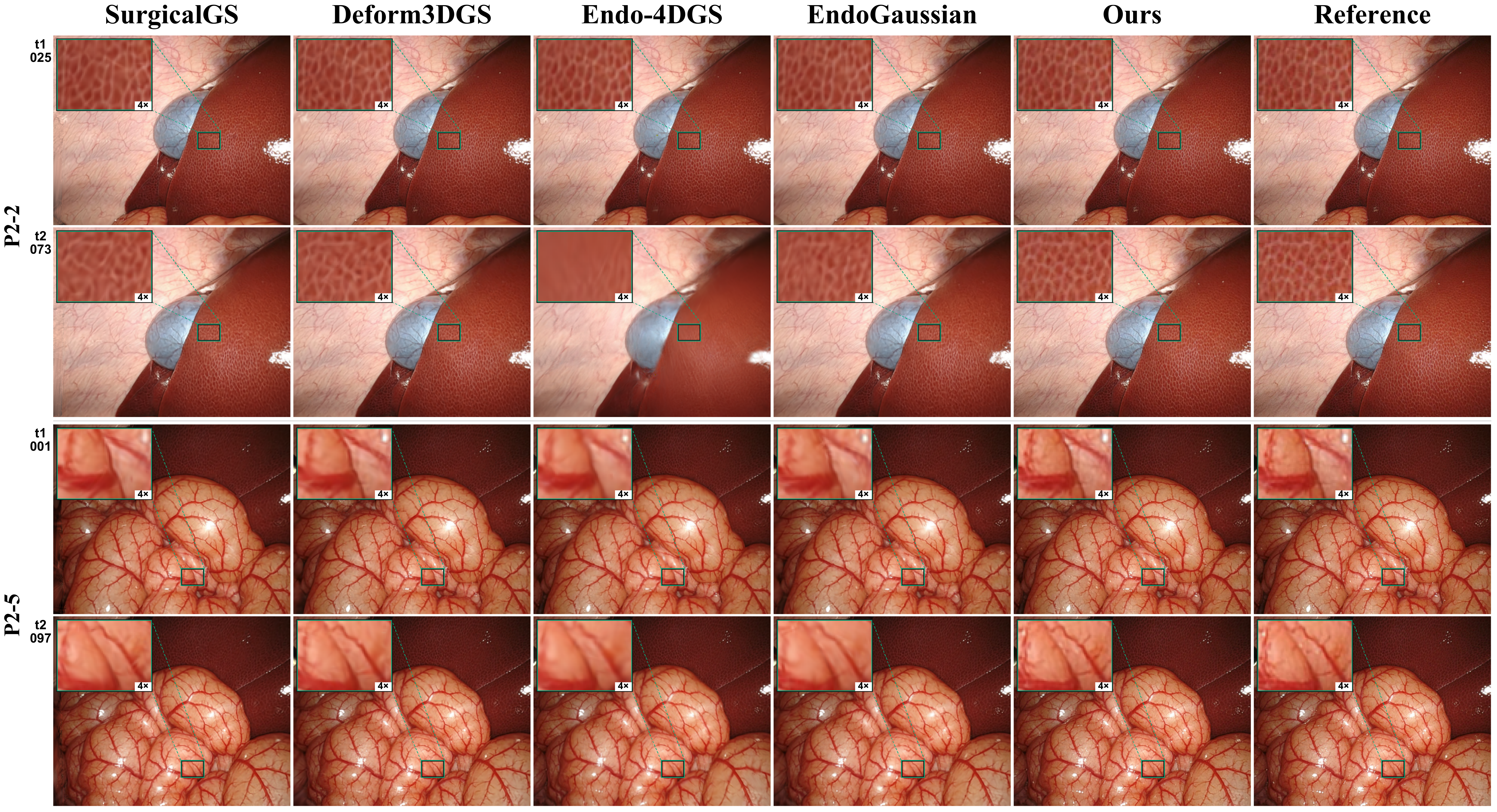}
    \caption{Held-out StereoMIS frames. The $4\times$ insets highlight reticular texture in P2-2 (top) and vessel detail in P2-5 (bottom). \method{} reproduces these structures more closely across the times shown.}
    \label{fig:qual_stereomis}
    \end{minipage}
\end{figure*}

\section{Experiments}
\label{sec:exp}

\subsection{Experimental setup}

We evaluate rendering fidelity on the six sequences used by Endo-PairGS~\cite{yu2026endopairgs}: EndoNeRF~\cite{wang2022neuralendonerf} cutting and pulling, and StereoMIS~\cite{hayoz2023learningstereomis} P1, P2-2, P2-5 and P3.
EndoNeRF captures tissue deformation with a near-static camera, whereas StereoMIS also includes camera motion and larger deformation.
We resize images to $640\times512$ and use the 7:1 training/test split of Endo-4DGS~\cite{huang2024endo4dgs}.
We fit each scene to the training images and evaluate renderings of held-out frames.

We report PSNR for photometric accuracy, SSIM~\cite{wang2004ssim} for structural similarity, and AlexNet LPIPS~\cite{zhang2018lpips} for perceptual similarity.
For the complete-pipeline comparison, Table~\ref{tab:sota} follows published full-image averaging, with instrument pixels zeroed in both images.
For component analysis, Table~\ref{tab:ablation} reports PSNR and SSIM computed over tissue pixels, retains masked-image LPIPS, and weights sequence means equally within each dataset.
Table~\ref{tab:sota} thus evaluates complete pipelines, while Table~\ref{tab:ablation} compares component configurations under a common tissue-focused protocol.

Metrics are averaged over test frames and then seeds, with five seeds for EndoNeRF and three for StereoMIS unless specified.
Paired changes are median differences between runs with the same seed.
Our matched-input variants share training-view initialisation, camera-extent settings, masked D-SSIM and depth inputs.
On StereoMIS, we use FoundationStereo~\cite{wen2025foundationstereo} depth priors, whereas the published methods retain their respective depth pipelines.

\noindent\textbf{Implementation.}
We use $F=8$ frequencies in $[1,16]$, RGB bound $\beta=0.3$, Mat\'ern length scale $\ell=0.2$ and $\lambda_{\mathrm{msp}}=1$.
Training runs for 6000 iterations, with temporal coefficients activated at $K_s=3000$ and cloning, splitting and pruning stopped at iteration 4000.
Calibration uses 16 temporally distributed training views from a base checkpoint at iteration 3000, targeting a median Laplacian-to-photometric image-gradient ratio of $0.25$ and yielding $\lambda_{\mathrm{lap}}\approx0.83$.
The detail weight increases linearly to this calibrated value from iteration 1000 over 1500 iterations.
Relative to the FDM coefficient learning rate, the learning rates for translation and colour coefficients use factors of $0.1$ and $31.25/16$, respectively, with both schedules restarted at $K_s$.
These settings are shared across sequences and seeds.

\subsection{Comparison with existing methods}

Table~\ref{tab:sota} compares \method{} with eight methods on the same sequences and split.
\method{} achieves the highest PSNR on all six sequences, the best or tied-best SSIM on all six, and the lowest LPIPS on five.
On cutting, it reaches 40.96\,dB PSNR and 0.019 LPIPS, and it leads each reported metric on the evaluated StereoMIS sequences.
On pulling, Endo-PairGS achieves the lowest LPIPS, whereas \method{} has higher PSNR and tied SSIM at the reported precision.

A matched-input comparison with SurgicalGS$^\dagger$ uses identical training inputs, initialisation and base reconstruction losses.
Across the sequences, \method{} achieves paired PSNR gains of $0.31$ to $0.66$\,dB and LPIPS reductions of $0.0035$--$0.0096$.
Sequence-level PSNR and LPIPS improve in every evaluated seed.
These controlled results complement the complete-pipeline comparison above.

\begin{table}[t]
\caption{\textbf{Component ablation.} A: temporal appearance. S: spectral motion refinement. L: multi-scale detail supervision. Scores weight sequence means equally. PSNR/SSIM use tissue pixels, while LPIPS uses masked images. Seeds: five for EndoNeRF (base: three), three for StereoMIS. Best results are bold.}
\centering
{\normalsize
\setlength{\tabcolsep}{2.2pt}
\renewcommand{\arraystretch}{1.15}
\begin{tabular}{@{}ccc ccc ccc@{}}
\toprule
& & & \multicolumn{3}{c}{EndoNeRF} & \multicolumn{3}{c}{StereoMIS} \\
\cmidrule(lr){4-6}\cmidrule(lr){7-9}
A & S & L & PSNR$\uparrow$ & SSIM$\uparrow$ & LPIPS$\downarrow$ & PSNR$\uparrow$ & SSIM$\uparrow$ & LPIPS$\downarrow$ \\
\midrule
$\times$ & $\times$ & $\times$ & 38.08 & 0.9537 & 0.0309 & 34.40 & 0.8959 & 0.0699 \\
$\times$ & $\checkmark$ & $\times$ & 38.12 & 0.9540 & 0.0302 & 34.44 & 0.8964 & 0.0686 \\
$\checkmark$ & $\checkmark$ & $\times$ & 38.40 & 0.9563 & 0.0280 & 34.74 & 0.9005 & 0.0643 \\
$\checkmark$ & $\checkmark$ & $\checkmark$ & \textbf{38.52} & \textbf{0.9569} & \textbf{0.0271} & \textbf{34.81} & \textbf{0.9015} & \textbf{0.0625} \\
\bottomrule
\end{tabular}}
\label{tab:ablation}
\end{table}

\noindent\textbf{Qualitative results.}
Figures~\ref{fig:qual_endonerf} and~\ref{fig:qual_stereomis} illustrate these rendering differences in held-out frames.
In EndoNeRF cutting, \method{} recovers clearer texture in the highlighted region, while the pulling examples show how it fits changing tissue appearance.
The StereoMIS insets highlight spatial detail: the reticular texture in P2-2 is more distinct, and the vessels in P2-5 more closely match the reference.

\subsection{Ablation study}

Table~\ref{tab:ablation} reports the incremental effects of temporal appearance (A), spectral motion refinement (S) and multi-scale detail supervision (L).
Density adaptation ends at iteration 4000 for all component configurations, including the configuration without A, S or L.
The motion-only variant adds the Fourier translation residual and Mat\'ern regularisation.
The A+S variant further adds the bounded colour residual, and the full model adds the Laplacian objective.
This cumulative design evaluates appearance with motion refinement active and detail supervision with both residuals active.

\noindent\textbf{Temporal appearance.}
With detail supervision disabled, adding the colour residual to the motion-only variant increases dataset-mean PSNR by $0.29$\,dB on EndoNeRF and $0.30$\,dB on StereoMIS.
These are the largest PSNR increments among the tested additions.
LPIPS decreases from $0.0302$ to $0.0280$ and from $0.0686$ to $0.0643$, respectively, and SSIM increases on both datasets.
These gains support explicit temporal appearance modeling alongside motion.

\noindent\textbf{Spectral motion refinement.}
Relative to the base, the motion-only variant improves dataset-mean PSNR by approximately $0.04$\,dB on both datasets.
SSIM increases and LPIPS decreases by $0.0007$ on EndoNeRF and $0.0013$ on StereoMIS.
To examine the Mat\'ern prior separately, we hold the Fourier translation representation and Gaussian counts fixed.
Adding the prior changes PSNR by $+0.005$\,dB on cutting and $-0.002$\,dB on pulling in the evaluated seed.
The prior provides frequency regularisation, with a small measured effect on rendering fidelity in this comparison.

\noindent\textbf{Multi-scale detail supervision.}
With both residuals active, removing the Laplacian objective reduces dataset-mean PSNR by $0.12$\,dB on EndoNeRF and $0.08$\,dB on StereoMIS.
Including this objective lowers LPIPS from $0.0280$ to $0.0271$ and from $0.0643$ to $0.0625$, respectively, and increases SSIM on both datasets.
Spatial detail supervision therefore further improves rendering when both temporal appearance and motion are modeled.

\noindent\textbf{Refinement with fixed Gaussian counts.}
To test refinement of existing primitives, we start the full model and motion-only variant from a shared iteration-3000 checkpoint and disable cloning, splitting and pruning in both runs.
Across three seeds, paired PSNR gains are $0.52$\,dB on cutting and $0.25$\,dB on pulling, with LPIPS reductions of $0.0024$ and $0.0028$.
Both variants refine the same initial scene with identical Gaussian counts.
Disabling density adaptation removes differences in primitive creation and removal, leaving the colour residual and detail objective as the changes under evaluation.
These results establish a joint benefit from temporal appearance and detail supervision without adding primitives.

\section{Conclusion}
\label{sec:conclusion}

We presented \method{} for dynamic endoscopic reconstruction with explicit temporal appearance and spectral motion refinement.
Its bounded colour residuals capture photometric variation, translation residuals refine trajectories under a spectral prior, and multi-scale supervision guides tissue-detail recovery.
Experiments on EndoNeRF and StereoMIS demonstrate state-of-the-art rendering quality, with appearance modeling providing the largest PSNR gain among the tested additions.
The joint benefit of appearance and detail supervision persists with fixed Gaussian counts, showing that refining the representation of existing primitives can improve rendering.

\clearpage

\end{document}